\documentclass[pdflatex,sn-mathphys-num]{sn-jnl}

\usepackage{graphicx}%
\usepackage{multirow}%
\usepackage{amsmath,amssymb,amsfonts}%
\usepackage{amsthm}%
\usepackage{mathrsfs}%
\usepackage[title]{appendix}%
\usepackage{xcolor}%
\usepackage{textcomp}%
\usepackage{manyfoot}%
\usepackage{booktabs}%
\usepackage{algorithm}%
\usepackage{algorithmicx}%
\usepackage{algpseudocode}%
\usepackage{listings}%
\usepackage{float} 
\usepackage{multirow} %
\usepackage{makecell}

\theoremstyle{thmstyleone}%
\theoremstyle{thmstyletwo}%

\theoremstyle{thmstylethree}%

\begin{document}

\title[Article Title]{A Survey on Vietnamese Document Analysis and Recognition: Challenges and Future Directions}


\author{\fnm{Anh} \sur{Le}}
\author{\fnm{Thanh} \sur{Lam}}
\author{\fnm{Dung} \sur{Nguyen}}






\abstract{Vietnamese document analysis and recognition (DAR) is a crucial field with applications in digitization, information retrieval, and automation. Despite advancements in OCR and NLP, Vietnamese text recognition faces unique challenges due to its complex diacritics, tonal variations, and lack of large-scale annotated datasets. Traditional OCR methods often struggle with real-world document variations, while deep learning approaches have shown promise but remain limited by data scarcity and generalization issues. Recently, large language models (LLMs) and vision-language models have demonstrated remarkable improvements in text recognition and document understanding, offering a new direction for Vietnamese DAR. However, challenges such as domain adaptation, multimodal learning, and computational efficiency persist. This survey provides a comprehensive review of existing techniques in Vietnamese document recognition, highlights key limitations, and explores how LLMs can revolutionize the field. We discuss future research directions, including dataset development, model optimization, and the integration of multimodal approaches for improved document intelligence. By addressing these gaps, we aim to foster advancements in Vietnamese DAR and encourage community-driven solutions.}




\maketitle

\section{Introduction}\label{sec1}
\subsection{Background for Document Analysis and Recognition}\label{subsec11}

\textbf{Document Analysis and Recognition} is a field of research focused on the automatic extraction, interpretation, and understanding of textual and structural information from documents. It encompasses various tasks, including Optical Character Recognition (OCR), handwriting recognition, layout analysis, and document classification. DAR plays a crucial role in digitization efforts, enabling automated data entry, archival preservation, and intelligent information retrieval applications. Traditional recognition techniques relied on handcrafted features and rule-based methods. However, with the advancement of deep learning and LLMs, modern approaches now leverage neural networks to achieve higher accuracy and adaptability across diverse document types and languages.

Among the many languages and scripts DAR systems must handle, Vietnamese presents unique challenges due to its complex diacritical system, tonal nature, and lack of large annotated datasets. Unlike Latin-based languages, Vietnamese words often contain multiple diacritics that significantly impact meaning and pronunciation, making accurate recognition more difficult.  Accurate text recognition must account for these marks, which are highly susceptible to recognition errors, mainly when dealing with noisy or degraded document images. Addressing these challenges is essential for successfully digitizing Vietnamese documents widely used in government, finance, and education and supporting AI-driven applications such as information retrieval and intelligent search systems. Vietnamese document analysis, therefore, requires specialized recognition models, robust preprocessing techniques, and the integration of advanced AI approaches to ensure reliable and efficient text extraction.

Additionally, Vietnamese words are not always separated by spaces in the same way as English, making word segmentation a significant issue. This problem is further compounded when working with historical documents, handwritten texts, or low-quality scans introducing distortions. Current recognition models trained on high-resource languages struggle with these linguistic intricacies, highlighting the need for tailored approaches in Vietnamese document recognition.

Vietnamese DAR primarily focuses on two main types of documents: \textbf{printed} and \textbf{handwritten} text. Each type presents unique challenges and requires specialized recognition techniques.
\begin{itemize} 
\item \textbf{Printed document recognition} involves extracting text from structured sources such as books, newspapers, identity cards, passports, and official records. These documents typically follow standard fonts and layouts, making them more amenable to OCR techniques. However, challenges arise due to noise, distortions, varying font styles, and complex backgrounds. The increasing digitization of official documents, such as identity cards, passports, and financial records, has led to a growing demand for robust OCR solutions. Studies such as \textit{"Text Recognition for Vietnamese Identity Cards Based on Deep Features Network"} \cite{van2021text} have addressed specific document types, but a generalizable, high-performance OCR system for diverse Vietnamese texts remains an open challenge.
\item On the other hand, \textbf{handwritten document recognition} is inherently more complex due to variations in writing styles, stroke connectivity, and inconsistencies in character formation. Unlike printed text, handwritten Vietnamese exhibits significant differences across individuals, making recognition highly challenging. Handwriting recognition is critical in document digitization, mainly for historical manuscripts, personal notes, and administrative forms. Research such as \textit{"A Database of Unconstrained Vietnamese Online Handwriting"} \cite{nguyen2018database} has highlighted the limitations of existing handwritten datasets and the need for improved models. The variability in handwriting requires sophisticated deep learning models, particularly sequence-based and attention-based architectures, to effectively recognize and interpret Vietnamese handwritten text.
\end{itemize}

In real-world applications, many documents combine printed and handwritten text, requiring OCR systems to handle both formats seamlessly. Examples include official documents such as identity cards, application forms, bank statements, and medical records, where structured printed text is supplemented with handwritten annotations, signatures, or form entries. Developing a unified DAR system that recognizes mixed-content documents remains an open challenge. Future improvements in vision-language models, self-supervised learning, and adaptive OCR pipelines will be crucial in bridging the gap between printed and handwritten text recognition, ensuring robust performance across diverse real-world applications.

Given these challenges, Vietnamese DAR requires more advanced research, such as multimodal approaches and the integration of LLMs, to enhance recognition accuracy and contextual understanding. Adopting vision-language models like TrOCR \cite{li2023trocr} and Donut \cite{kim2022ocr}, dictionary-guided text recognition \cite{nguyen2021dictionary} has shown promise in addressing Vietnamese-specific OCR challenges. However, significant work remains in dataset expansion, model optimization, and developing specialized benchmarks to push the field forward.

\subsection{Scope and Contributions}\label{subsec12}
This survey aims to provide a comprehensive overview of Vietnamese document analysis and recognition, highlighting the progress made in the field, identifying existing gaps, and suggesting future research directions. Specifically, our contributions include:
\begin{itemize}
\item \textbf{Comprehensive Review}: We conduct a detailed survey of DAR techniques tailored to the Vietnamese language. This includes traditional OCR methods, deep learning-based approaches, and emerging LLM-driven solutions.

\item \textbf{Dataset Analysis}: We analyze existing Vietnamese OCR datasets, including printed, handwritten, and scene text corpora, and discuss their limitations regarding size, diversity, and annotation quality.

\item \textbf{Evaluation of LLM-based Approaches}: We explore the role of Large Language Models in Vietnamese document analysis, discussing their strengths, weaknesses, and potential for improving recognition accuracy and document understanding.

\item \textbf{Challenges and Open Problems}: We identify significant challenges in Vietnamese OCR, including linguistic complexities, dataset scarcity, and model limitations, and highlight areas where further research is needed.

\item \textbf{Future Research Directions}: We propose future directions in Vietnamese document analysis, emphasizing multimodal learning, dataset expansion, transfer learning, and computational efficiency.
\end{itemize}

By addressing these key areas, this survey aims to serve as a valuable resource for researchers and practitioners working on Vietnamese document analysis and encourage further advancements in the field.

\section{Vietnamese Document Analysis and Recognition: Current Approaches}\label{sec2}
\subsection{Traditional DAR Methods}\label{subsec21}
Traditional Document Analysis and Recognition (DAR) methodologies predominantly employ rule-based techniques, template matching, and handcrafted feature extraction to recognize printed text. Extensive research has been conducted on integrating handcrafted feature extraction with classical algorithms such as Support Vector Machines (SVM), K-nearest neighbors (KNN), and Random Forests for languages like English \cite{bunke1995off, nasien2010support}. However, due to limited research resources, few studies are focusing on traditional methods for Vietnamese isolated character recognition. Pham et al. \cite{phuong2008speeding} proposed an efficient model for isolated Vietnamese handwritten character recognition, utilizing SVMs and statistical feature extraction methods. By implementing feature dimension reduction and the Reduced Set Method (RSM), their model achieved over 90\% accuracy while enhancing recognition speed by ninefold compared to the original approach. Experimental results on 52,485 character samples demonstrated the model's effectiveness; however, further research is required to extend this approach to full word and sentence recognition.

Several open-source OCR engines, such as Tesseract, have been adapted for Vietnamese text recognition. Although Tesseract supports Vietnamese, its accuracy remains limited due to diacritic handling and document noise challenges. Phan et al. \cite{nga2017vietnamese} demonstrate how traditional OCR methods fail in complex backgrounds, requiring additional preprocessing and enhancement techniques. They present a method for extracting Vietnamese text from scanned book cover images. The process involves image preprocessing to enhance quality, text region localization using techniques like the Sobel filter, and OCR with Tesseract, explicitly trained for Vietnamese. Post-processing includes dictionary filtering to refine the extracted text, achieving promising results on their dataset.

On the other hand, Phan et al. \cite{van2016nom} developed a system for digitizing historical Vietnamese documents written in the Nôm script. The system employs image binarization, character segmentation using recursive X–Y cuts and Voronoi diagrams, and character recognition through k–d tree classification and modified quadratic discriminant functions. It also provides a user interface for result verification and correction, enhancing the accuracy of the digitization process.

\subsection{Deep Learning-Based DAR Methods}\label{subsec22}

Deep learning has significantly advanced OCR for Vietnamese text, addressing challenges such as complex diacritics and varied document formats. However, the diversity of in-house datasets and the lack of standardized evaluation metrics pose challenges for comparative analysis.

The Vietnamese Online Handwritten Text Recognition (VOHTR) competition, organized during the International Conference on Frontiers in Handwriting Recognition (ICFHR) 2018 \cite{nguyen2018icfhr}, aimed to evaluate and compare online handwritten text recognition systems using the VNOnDB database \cite{nguyen2018database}. Participants tackled word, line, and paragraph tasks, addressing challenges such as delayed strokes caused by diacritic marks.

The methods used in the VOHTR 2018 competition include:
\begin{itemize}
\item \textbf{Google Team:} Applied preprocessing techniques such as scaling, normalization, resampling, and Bezier curve representation to the ink data. The processed data was then fed into multiple Bidirectional Long Short-Term Memory (BLSTM) layers, with post-processing using n-gram language models at both character and word levels, trained on their proprietary corpus.
\item \textbf{IVTOV Team:} Implemented line segmentation prior to preprocessing. Extracted online features were used to train a recognition network comprising two BLSTM layers, each with 100 cells. Post-processing involves applying dictionary constraints to the output sequences.
\item \textbf{MyScript System:} Preprocessed digital ink by normalizing with Bezier approximation and correcting slope and slant. The system included two recognizers: a feed-forward network for predicting characters from segmented candidates and a BLSTM network for predicting output text without segmentation. Post-processing utilized a syllable-based n-gram language model trained on a large corpus, including the Vietnamese Treebank (VTB) \cite{phuong2009building} and additional lexica.
\end{itemize}

Moreover, the authors of the VNOnDB dataset \cite{nguyen2018database} introduced a BLSTM-based network that leverages both online and offline extracted features to recognize text at the line and paragraph levels. Subsequently, Le et al. \cite{le2018recognizing} applied an attention-based encoder-decoder (AED) model, incorporating standard CNN and BLSTM encoders along with an attention-enabled LSTM decoder, for offline text recognition. In their approach, online handwritten texts are first transformed into images, which are then processed by the AED model to generate OCR outputs. This model was evaluated exclusively on the word-level VNOnDB dataset. Later, Le et al. \cite{le2020end} proposed an enhanced end-to-end AED model featuring a DenseNet encoder, and validated its performance on both the word-level and line-level VNOnDB datasets. Notably, none of these three studies employed language models during the post-processing phase.

These methods were evaluated on the VNOnDB datasets at word, line, and paragraph levels (see Table \ref{tab:vnm_onl_ocr_studies}). The competition played a pivotal role in advancing research in Vietnamese handwriting recognition by providing a benchmark for comparing different approaches and encouraging the development of more accurate and efficient OCR systems.

\begin{table}[!htbp]
\caption{Methods for The Vietnamese Online Handwritten Text Recognition Competition}
\label{tab:vnm_onl_ocr_studies}
\centering
\begin{tabular}{@{}p{1.2cm} p{2.2cm} p{1.8cm} p{1.5cm} p{1cm} p{2.8cm}@{}}
\toprule
\textbf{Level} & \textbf{Method} & \textbf{Text line extraction preprocessing} & \textbf{Language model} & \textbf{Corpus} & \textbf{Recognizer} \\
\midrule
\multirow{5}{*}{Word} & GoogleTask1 & No & Character-based and word-based n-gram & Other & Multi LSTM layers + CTC \\
& IVTOVTask1 & Yes & Dictionary & VTB & 2 BLSTM layers + CTC \\
& MyScriptTask1\_1 & No & Syllable-based unigram & VTB + others & Segmentation + Feedforward Neural Network (FNN) \& BLSTM + CTC \\
& AED model with standard CNN and BLSTM encoders \cite{le2018recognizing} & No & None & None & Standard CNN and BLSTM encoders and attention-based LSTM decoder \\
& AED model with DenseNet encoder \cite{le2020end} & No & None & None & DenseNet encoder and attention-based LSTM decoder \\
\midrule
\multirow{6}{*}{Line} & GoogleTask2 & No & Character-based and word-based n-gram & Other & Multi LSTM layers + CTC \\
& IVTOVTask2 & Yes & Dictionary & VTB & 2 BLSTM layers + CTC \\
& MyScriptTask2\_1 & No & Syllable-based trigram & VTB & Segmentation + FNN \& BLSTM + CTC \\
& MyScriptTask2\_2 & No & Syllable-based trigram & VTB + others & Segmentation + FNN \& BLSTM + CTC \\
& Two-layered BLSTM model using extracted features \cite{nguyen2018database} & Yes & None & None & BLSTM with two hidden layers \\
& AED model with DenseNet encoder \cite{le2020end} & No & None & None & DenseNet encoder and attention-based LSTM decoder \\
\midrule
\multirow{5}{*}{Paragraph} & IVTOVTask3 & Yes & Dictionary & VTB & 2 BLSTM layers + CTC \\
& MyScriptTask3\_1 & Yes & Word-based trigram & VTB & Segmentation + FNN \& BLSTM + CTC \\
& MyScriptTask3\_2 & Yes & Syllable-based trigram & VTB + others & Segmentation + FNN \& BLSTM + CTC \\
& MyScriptTask3\_3 & Yes & Word-based trigram & VTB & Segmentation + FNN \& BLSTM + CTC \\
& Two-layered BLSTM model using extracted features \cite{nguyen2018database} & Yes & None & None & BLSTM with two hidden layers \\
\botrule
\end{tabular}
\end{table}

In addition, the Mobile-Captured Image Document Recognition for Vietnamese Receipts (MC-OCR) Challenge \cite{vu2021mc} was organized in conjunction with the RIVF 2021 conference to advance the field of OCR for Vietnamese receipts captured via mobile devices. The competition aimed to address the complexities associated with such images, including issues like crumpled receipts, blurring, and varied lighting conditions. There are 2 tasks: 
\begin{enumerate}
\item Receipt Quality Assessment (IQA): Evaluate the readability of receipt images based on the clarity of extracted text.
\item Key Information Extraction (KIE): Accurately extract specific fields from receipts, namely "SELLER," "SELLER ADDRESS," "TIMESTAMP," and "TOTAL COST."
\end{enumerate}

The dataset comprised over 2,000 images of Vietnamese receipts captured using mobile devices. These images were contributed by more than 50 volunteers who photographed their own or acquaintances' receipts. Human annotators then labeled each text line within the receipts, indicating the corresponding text and assessing the image quality based on the number of clear text lines.

Six teams submitted detailed reports on their methodologies and results. Table \ref{tab:vnm_mobile_ocr_studies} presents a summary of their approaches and performance.

\begin{table}[!htbp]
\caption{Participants for The Mobile-Captured Image Document Recognition for Vietnamese Receipts Challenge}
\label{tab:vnm_mobile_ocr_studies}
\centering
\begin{tabular}{@{}p{2.5cm} p{7cm} p{1.2cm} p{1cm}@{}}
\toprule
\textbf{Team} & \textbf{Methodology} & \textbf{Task 1 (RMSE)} & \textbf{Task 2 (CER)} \\
\midrule
DataMining VC \cite{nguyenmc2021mc} & \makecell[l]{Task 1: Patch sifting.\\ Task 2: Implemented an information detection step\\ using Yolov5, involving text-block localization and\\ classification, followed by OCR using VietOCR.} & 0.15 & 0.22 \\

SDSV AICR \cite{nguyendc2021mc} & \makecell[l]{Task 1: Explored three methods: detecting image blur,\\ averaging PaddleOCR confidence scores, and a regression\\ model using depth-wise separable convolution (DSC).\\ Task 2: Pipeline with PaddleOCR text detection,\\ MobileNet v3 rotation correction, VietOCR text\\ recognition, and key information extraction using\\ GCN-based model by PICK.} & 0.12 & 0.23 \\

SUN-AI \cite{nguyenvh2021mc} & \makecell[l]{Task 1: Multilayer perceptron-based regression model\\ using the 100 lowest VietOCR confidence scores\\ after segmentation by CRAFT.\\ Task 2: SVM for "TOTAL COST," PhoBERT for\\ "SELLER" and "ADDRESS," and rule-based\\ "TIMESTAMP" extraction.} & 0.15 & 0.26 \\

Tung-Nguyen \cite{bui2021mc} & \makecell[l]{Task 2: Faster R-CNN for information location\\ detection, and TransformerOCR (CNN + Transformer)\\ for text recognition.} & N/A & 0.32 \\

UIT CS AIClub \cite{le2021mc} & \makecell[l]{Task 1: EfficientNet-based regression model.\\ Task 2: Preprocessing, text detection using PAN,\\ VietOCR text recognition, and rule-based structured\\ information extraction.} & 0.10 & 0.30 \\

BK OCR \cite{hieu2021mc} & \makecell[l]{Task 1: VGG-16 regression model.\\ Task 2: CRAFT text detection, VietOCR text\\ recognition, and rule-based information extraction.} & 0.11 & 0.39 \\
\botrule
\end{tabular}
\end{table}

\begin{itemize}
\item For Task 1, teams predominantly employed regression-based models utilizing features derived from OCR confidence scores, image quality metrics (e.g., blur detection), and convolutional neural networks (CNNs) such as EfficientNet \cite{tan2019efficientnet} and VGG-16 \cite{simonyan2014very}.
\item For Task 2, teams implemented multi-stage OCR pipelines comprising:
\begin{itemize}
    \item \textbf{Text Detection:} Techniques including Yolov5 \cite{ultralytics2021yolov5}, CRAFT \cite{baek2019character}, PAN \cite{wang2019efficient}, Faster R-CNN \cite{ren2015faster}, and PaddleOCR \cite{du2020pp}.
    \item \textbf{Rotation Correction and Alignment:} CNN-based models such as MobileNet v3 \cite{howard2019searching}.
    \item \textbf{Text Recognition:} Transformer-based OCR models (VietOCR \cite{vietocr2020}, TransformerOCR).
    \item \textbf{Information Extraction:} Methods employing graph convolutional networks (GCN) like PICK \cite{yu2021pick}, SVM, PhoBERT \cite{nguyen2020phobert}, and rule-based systems.
\end{itemize}
\end{itemize}

The top-performing approach for Task 1 utilized CNN-based regression models, such as EfficientNet, trained on extracted text-image features to predict image quality, achieving RMSE scores as low as 0.10 (UIT CS AIClub team). For Task 2, the DataMining VC team achieved the best character error rate (CER) of 0.22 by introducing a dedicated information localization/classification stage prior to OCR.

Overall, the top-performing methods underscored the efficacy of integrating CNN and Transformer architectures within structured pipelines, significantly improving OCR accuracy and document quality assessment. The variety of methodologies utilized by leading teams provides valuable insights into effective OCR strategies tailored specifically to Vietnamese receipts. 

Both competitions have played pivotal roles in unifying datasets and evaluation metrics for Vietnamese DAR. They have attracted research attention by providing standardized benchmarks and fostering advancements in Vietnamese handwriting recognition and OCR technologies. These efforts have facilitated the comparison of different methodologies and encouraged the development of more accurate and efficient recognition systems tailored to the Vietnamese language.

Other than the above method, recent research has significantly advanced Vietnamese text recognition in various contexts. Huynh et al. (2023) \cite{huynh2023scene} introduced a method for recognizing Vietnamese text in scene images combining character sequence prediction, context processing, and iterative correction, achieving word-level accuracies of 81.87\% on the VinText dataset and 82.56\% on the VnSceneText dataset. In a different application, Phan et al. (2021) \cite{van2021text} presented a technique specifically designed for Vietnamese identity cards using deep feature networks, which attained over 96.7\% character-level accuracy and 89.7\% word-level accuracy. Expanding the scope further, Pham et al. (2024) \cite{pham2024viocrvqanovelbenchmarkdataset} created ViOCRVQA, a benchmark dataset for Visual Question Answering (VQA) involving Vietnamese text, comprising over 28,000 images and 120,000 question-answer pairs, with performance metrics reported as an Exact Match (EM) of 0.4116 and an F1-score of 0.6990. Additionally, Dinh et al. (2023) \cite{dinh2023vietnamese} addressed the challenge of digitizing handwritten medical records through a pipeline that incorporates text area detection, enhancement, transcription, and auto-correction, achieving a Character Error Rate (CER) of 2\% and a Word Error Rate (WER) of 12\%.

\subsubsection{Vietnamese OCR and Handwriting Recognition Datasets}\label{subsubsec221}
The development of Vietnamese OCR systems often relies on in-house datasets tailored to specific applications, such as identity cards, receipts, or medical records. This diversity leads to dataset size variations, image quality, and text complexities. Consequently, a lack of unified evaluation metrics makes it challenging to compare the performance of different OCR models across studies. Standardizing datasets and evaluation protocols is essential for advancing the field and facilitating meaningful comparisons.

Efforts like creating benchmark datasets such as ViOCRVQA and ViTextVQA \cite{doan2024vintern} are steps toward addressing these challenges by providing standardized resources for training and evaluating Vietnamese OCR models. However, further collaboration and consensus within the research community are necessary to establish unified metrics and datasets.

Vietnamese OCR and handwriting recognition have seen significant advancements, supported by various datasets tailored to different applications. Table \ref{tab:vnm_ocr_datasets} is a survey of notable datasets in this domain.

\begin{table}[!htbp]
\caption{Summary Table of Vietnamese OCR and Handwriting Recognition Datasets}\label{tab:vnm_ocr_datasets}%
\centering
\begin{tabular}{@{}p{3cm} p{3cm} p{1.5cm} p{2.5cm} p{1cm}@{}}
\toprule
\textbf{Dataset Name} & \textbf{Description} & \textbf{Size} & \textbf{Annotations} & \textbf{Source} \\
\midrule
4,995 Vietnamese OCR Images Dataset & Images of natural scenes, internet images, and documents with annotations. & 4,995 images & Line-level and column-level bounding boxes with text transcriptions & GitHub \cite{nexdata} \\
Vietnamese Handwriting Recognition Dataset by Cinnamon AI & Handwritten text images with labels in JSON format. & 1,838 images & Image labels in JSON format & GitHub \cite{cinnamonaidata} \\
Vietnamese Handwritten OCR Dataset on Kaggle & Images of Vietnamese words with corresponding labels. & Over 110,000 images & Image arrays and labels in CSV format & Kaggle \cite{kaggledata} \\
Vietnamese OCR Image Corpus by DataoceanAI & Printed images covering common daily scenarios with annotations. & 5,006 images & Line-level bounding boxes and text transcriptions & Dataocean AI \cite{dataoceanaidata} \\
Vietnamese Online Handwriting Database (HANDS-VNOnDB) & Paragraphs of handwritten text with lines, strokes, and characters. & 1,146 paragraphs & Lines, strokes, and character annotations & TC-11 \cite{nguyen2018database} \\
Unconstrained Handwriting Image Recognition Dataset (UIT-HWDB) & Synthetic handwriting images for evaluating recognition methods. & Not specified & Not specified & RIVF \cite{nguyen2022data} \\
Visual Question Answering Dataset with Vietnamese Text (ViOCRVQA) & Images with text and corresponding question-answer pairs. & Over 28,000 images & Question-answer pairs related to text in images & arXiv \cite{pham2024viocrvqanovelbenchmarkdataset} \\
\botrule
\end{tabular}
\end{table}

\subsection{Vietnamese OCR Post-Processing Approaches}

OCR post-processing is crucial for enhancing the accuracy of OCR-generated texts by detecting and correcting errors. Various methods have been proposed, including corpus-based language models, machine learning techniques, evolutionary algorithms, large language models, and neural machine translation approaches \cite{nguyen2019ocr, nguyen2021ocr, nguyen2023efficient, do2024reference, hoang2012unsupervised}.

The main categories of OCR post-processing approaches for Vietnamese are as follows:
\begin{itemize}
\item Word-Level n-gram Models with Error Modeling  \cite{nguyen2019ocr}: This method, called statistical language model (SLM), utilizes word-level n-gram models combined with an error model that learns correction character edits from aligned ground truth and OCR training texts to detect and correct OCR errors.

\item Neural Machine Translation (NMT) with Bidirectional LSTM \cite{nguyen2021ocr}: A word-level NMT model based on BLSTM networks is employed to correct OCR errors, leveraging the sequential nature of language for improved accuracy.

\item Unsupervised OCR Error Correction Using Hill Climbing (HC) Algorithm \cite{nguyen2023efficient}: This approach proposes an unsupervised OCR error correction model based on an adapted HC algorithm, which explores and selects correction candidates using random character edits extracted from original ground truth texts without prior knowledge of OCR error characteristics.

\item Reference-Based Post-OCR Processing with LLMs \cite{do2024reference}: This method utilizes content-focused ebooks as a reference base, supported by LLMs, to correct imperfect OCR-generated text. It addresses challenges such as missing characters, words, and disordered sequences, particularly in historical documents with diacritic languages like Vietnamese.

\item Weighting-Based Model with Contextual Corrector \cite{hoang2012unsupervised}: This approach employs two correction models: one based on syllable similarity and frequency and another using language modeling based on perplexity scores. It implements a depth-first traversal algorithm to examine all candidate combinations, which can be computationally intensive due to the high number of syllables.
\end{itemize}

While the two OCR post-processing studies mentioned above \cite{do2024reference, hoang2012unsupervised} evaluated their methods on own datasets, the remaining works assessed their approaches using OCR-generated text from the VNOnDB database, produced by the AED model \cite{le2020end}. These post-processing techniques contributed to further enhancing OCR output quality by reducing the CER and WER error rates as shown in Table \ref{tab:vnm_postocr}. The diversity of approaches highlights the ongoing efforts to improve OCR post-processing for Vietnamese, addressing the unique challenges posed by the language's diacritic system and complex character structures.

\begin{table}[!htbp]
\caption{The Performance of OCR Post-processing Approaches on The VNOnDB-OCR Dataset at Line Level}\label{tab:vnm_postocr}%
\centering
\begin{tabular}{@{}p{2.5cm} p{2.5cm} p{1cm} p{1cm} p{1cm}@{}}
\toprule
\textbf{Method} & \textbf{Language model} & \textbf{Corpus} & \textbf{CER (\%)} & \textbf{WER (\%)} \\
\midrule
GoogleTask2 & Character-based and word-based n-gram & Other & 6.86 & 19.00 \\
IVTOVTask2 & Dictionary & VTB & 3.24 & 14.11 \\
MyScriptTask2\_1 & Syllable-based trigram & VTB & 1.02 & 2.02 \\
MyScriptTask2\_2 & Syllable-based trigram & VTB + others & 1.57 & 4.02 \\
Two-layered BLSTM model using extracted features \cite{nguyen2018database} & None & None & 7.17 & NA \\
AED model with DenseNet encoder \cite{le2020end} & None & None & 4.67 & 13.33 \\
SLM model \cite{nguyen2019ocr} & Syllable-based n-gram and character edits & VTB & 4.17 & 9.82 \\
BLSTM-based NMT model \cite{nguyen2021ocr} & Syllable-based unigram and BLSTM & VTB & NA & 11.5 \\
HC-based unsupervised model \cite{nguyen2023efficient} & Syllable-based n-gram and correction patterns & VTB & 4.13 & 9.47 \\
\botrule
\end{tabular}
\end{table}

\section{Challenges in Vietnamese Document Recognition}\label{sec3}

Vietnamese Document Recognition faces several unique challenges due to linguistic complexities, dataset limitations, model constraints, and the limited number of researchers and research communities dedicated to this field.

\subsection{Linguistic Challenges}\label{subsec31}
\textbf{Vietnamese Diacritics and Their Impact on OCR Accuracy:} Vietnamese utilizes a Latin-based script enriched with numerous diacritical marks that denote tones and vowel distinctions. These diacritics are essential for conveying meaning, as words with identical spellings can have different interpretations based on tonal variations. For instance, the words "ma" (ghost), "má" (mother), and "mà" (but) differ solely in their diacritical marks. OCR systems often struggle to accurately recognize these diacritics, especially when they appear both above and below characters, leading to misinterpretations of the text. This issue is particularly pronounced in scene text recognition, where complex backgrounds and varying fonts further complicate diacritic detection.

\textbf{Word Segmentation and Tonal Ambiguities:} Unlike languages that use spaces to delineate words, Vietnamese often combines multiple syllables to form words, and spaces may not always indicate word boundaries. This characteristic poses significant challenges for word segmentation algorithms, which are crucial for accurate text analysis and recognition. Incorrect segmentation can lead to misinterpretation of the text, affecting downstream tasks such as translation and information extraction. Additionally, the tonal nature of Vietnamese introduces ambiguities, as the same sequence of letters can represent different words depending on the tone, further complicating the recognition process.

\subsection{Dataset Limitations}\label{subsec32}
\textbf{Lack of Large-Scale Annotated Datasets for Vietnamese OCR:} The development of robust OCR systems requires extensive annotated datasets that capture the linguistic and visual diversity of the target language. However, Vietnamese OCR research suffers from a scarcity of such datasets. While some efforts have been made to create datasets like ViOCRVQA, which contains over 28,000 images with text and corresponding question-answer pairs, the overall availability of large-scale, high-quality annotated datasets remains limited. This shortage hampers the training and evaluation of OCR models, restricting advancements in the field.

\textbf{Need for Diverse Document Types:} Effective OCR systems must handle a variety of document types, including handwritten notes, printed materials, scanned documents, and images captured under different conditions. The existing Vietnamese datasets often lack this diversity, focusing predominantly on specific document types or conditions. For example, the VNDoc dataset \cite{le2023dataset} includes 226 documents scanned from mobile phones and scanners, covering categories like legal documents, invoices, and resumes. However, the limited scope of such datasets may not fully represent the range of documents encountered in real-world scenarios, affecting the generalizability of OCR models.

\subsection{Model Limitations}\label{subsec33}
\textbf{Poor Generalization Across Different Document Domains:} OCR models trained on specific document types or conditions may perform poorly when applied to different domains. For instance, a model trained exclusively on printed text may struggle with handwritten documents or texts captured in varying lighting conditions. This lack of generalization is a significant hurdle, as it necessitates the development of versatile models capable of adapting to diverse document characteristics.

\textbf{Lack of LLMs Specialized for Vietnamese Document Analysis:} While LLMs have shown promise in various natural language processing tasks, their application to Vietnamese document analysis is limited due to the scarcity of models specifically trained on Vietnamese data. This gap restricts the ability to leverage advanced language understanding capabilities for tasks such as text correction, information extraction, and semantic analysis within Vietnamese documents.

\subsection{Research Community Limitations}\label{subsec34}
\textbf{Limited Number of Researchers in the Field:} The field of Vietnamese Document Recognition is relatively niche, with a limited number of researchers focusing on its unique challenges. This scarcity can be attributed to factors such as the complexity of the Vietnamese language, the lack of comprehensive datasets, and limited funding opportunities. The small research community results in fewer collaborative efforts, slower progress, and a reduced number of publications, which in turn affects the visibility and advancement of the field.

\textbf{Lack of a Unified Research Community and Direction:} The absence of a cohesive research community dedicated to Vietnamese Document Recognition leads to fragmented efforts and a lack of standardized benchmarks and evaluation metrics. Without a unified direction, researchers may duplicate efforts, and advancements may not be effectively shared or built upon. Establishing a dedicated research community or consortium could facilitate collaboration, resource sharing, and the development of standardized datasets and evaluation protocols, thereby accelerating progress in the field.

Addressing these challenges requires concerted efforts in developing comprehensive datasets, refining linguistic processing techniques, creating adaptable models tailored to the nuances of the Vietnamese language and its diverse document types, and fostering a collaborative research community to provide direction and support for advancements in Vietnamese Document Recognition.

\section{The Role of LLMs in Vietnamese Document Analysis}\label{sec4}

LLMs have significantly advanced natural language processing tasks, including document analysis. Their application to Vietnamese document analysis offers promising avenues for improving text recognition and understanding.

\subsection{LLMs for Text Recognition and Understanding}\label{subsec41}

\textbf{Multimodal LLMs Enhancing Vietnamese OCR:} Multimodal LLMs, which process both textual and visual information, have shown potential to improve OCR for Vietnamese. For instance, Vintern-1B \cite{doan2024vintern} integrates the Qwen2-0.5B-Instruct language model with the InternViT-300M-448px visual model, optimizing it for tasks like OCR and document extraction in Vietnamese contexts. Fine-tuned on over 3 million image-question-answer pairs, Vintern-1B demonstrates robust performance across benchmarks such as OpenViVQA and ViTextVQA.

\textbf{Attention-Based Methods for Layout Understanding:} Attention mechanisms in LLMs enable models to focus on relevant parts of the input data, enhancing layout understanding in documents. This capability is crucial for accurately interpreting complex document structures, such as tables and forms, which are common in Vietnamese documents. By effectively capturing spatial relationships and formatting nuances, attention-based methods contribute to more accurate text extraction and comprehension.

\subsection{Challenges in Applying LLMs to Vietnamese OCR}\label{subsec42}
 
\textbf{Lack of Pretraining Data for Vietnamese Document Images:} The effectiveness of LLMs heavily relies on the availability of extensive and diverse pretraining data. However, there is a notable scarcity of large-scale annotated datasets for Vietnamese document images. This limitation hinders the development and fine-tuning of LLMs tailored to Vietnamese OCR tasks, impacting their performance and generalization capabilities.

\textbf{High Computational Costs and Data Efficiency Issues:} Training and deploying LLMs require substantial computational resources, which can be a barrier, especially in regions with limited infrastructure. Additionally, LLMs often demand large amounts of data to achieve optimal performance, posing challenges in contexts where data availability is constrained. Efforts to develop more efficient models, such as Vintern-1B, aim to mitigate these issues by creating models that are both effective and resource-conscious.

Addressing these challenges necessitates collaborative efforts to compile comprehensive datasets, optimize model architectures for efficiency, and develop LLMs specifically designed for Vietnamese document analysis. Such initiatives will enhance the accuracy and applicability of OCR systems in processing Vietnamese texts.

\section{Future Research Directions}\label{sec5}
\subsection{Developing High-Quality Vietnamese Datasets}\label{subsec51}

\textbf{Crowdsourcing and Synthetic Data Generation:} The scarcity of large-scale, annotated datasets for Vietnamese OCR poses significant challenges. To address this, researchers have explored synthetic data generation techniques. For instance, the VietNamese-OCR-DataGenerator \cite{VnOCRDataGen} is a tool designed to create synthetic text images for OCR training, supporting non-Latin scripts and various fonts. Similarly, the TextRecognitionDataGenerator \cite{TextRecogDataGen} facilitates the generation of text image samples, aiding in the training of OCR systems for diverse languages, including Vietnamese. These tools enable the creation of diverse and extensive datasets, which are crucial for training robust OCR models.

\textbf{Benchmarking Across Diverse Vietnamese Document Types:} The development of comprehensive benchmarks is essential for evaluating OCR systems across various document types. The ViOCRVQA dataset, for example, offers a collection of over 28,000 images with corresponding question-answer pairs, facilitating the assessment of visual question-answering systems in Vietnamese. Additionally, the UIT-HWDB dataset provides a benchmark for evaluating unconstrained handwriting image recognition in Vietnamese, addressing the need for diverse handwriting samples. These benchmarks are vital for assessing model performance and guiding future research.

\textbf{Collaborative Initiatives Between Academia and Industry:} The creation of large-scale, high-quality datasets for Vietnamese OCR can be significantly accelerated through collaborations between academic institutions and industry partners. Such partnerships enable academia to access proprietary datasets and resources while the industry benefits from academic expertise and research outcomes. For instance, collaborations have led to the development of datasets like ViTextVQA, which combines academic research with industry resources to create a comprehensive dataset for Vietnamese visual question answering. These joint efforts facilitate the development of cutting-edge AI technologies and provide researchers with real-world applications to validate their findings.

\subsection{Improving Model Robustness and Efficiency}\label{subsec52}

\textbf{Fine-Tuning Vision-Language Models for Vietnamese OCR:} Adapting vision-language models to the Vietnamese context is crucial for enhancing OCR performance. Fine-tuning pre-trained models on Vietnamese datasets can improve their ability to recognize and interpret Vietnamese text accurately. This approach leverages existing model architectures while tailoring them to the linguistic nuances of Vietnamese.

\textbf{Exploring Self-Supervised and Few-Shot Learning Techniques:} Self-supervised learning has shown promise in improving model robustness, especially in scenarios with limited labeled data. For instance, self-supervised learning techniques have been applied to enhance the performance of models in few-shot learning settings, demonstrating improved accuracy and generalization. Applying these techniques to Vietnamese OCR could mitigate the challenges posed by data scarcity and improve model efficiency.

\subsection{Expanding Multimodal Document Understanding}\label{subsec53}

\textbf{Integrating OCR with Document Layout and Content Extraction:} A comprehensive understanding of documents requires not only text recognition but also the analysis of document layout and content. Integrating OCR with layout analysis enables the extraction of structured information, such as tables and forms, which are prevalent in Vietnamese documents. This integration facilitates more accurate and context-aware information retrieval.

\textbf{End-to-end Vietnamese Document AI Systems:} Developing end-to-end systems that encompass OCR, layout analysis, and content extraction is a promising direction for Vietnamese document analysis. Such systems would streamline the processing of Vietnamese documents, enabling applications like automated data entry, document classification, and information extraction. Advancements in multimodal models and the availability of comprehensive datasets are pivotal for realizing these systems.

Addressing these research directions will significantly advance the field of Vietnamese document analysis, leading to more accurate and efficient processing of Vietnamese texts across various applications.

\section{Conclusion}\label{sec6}
 
This paper examined key challenges in Vietnamese document recognition, highlighting linguistic complexities such as diacritics, tonal variations, and word segmentation difficulties that hinder OCR accuracy. Additional limitations include the scarcity of diverse, large-scale annotated datasets, poor model generalization across different document domains, and the absence of specialized large language models tailored for Vietnamese.

Emerging multimodal LLMs, like Vintern-1B, offer promising improvements in text recognition, comprehension, and document layout understanding through attention-based mechanisms. However, applying LLMs to Vietnamese OCR faces barriers such as limited pretraining data and high computational demands. Future progress relies on creating comprehensive datasets, exemplified by ViOCRVQA, and strengthened collaboration between academia and industry to standardize benchmarks, improve resource sharing, and advance the field efficiently.

In conclusion, while significant challenges persist in Vietnamese document recognition, the strategic application of LLMs and collaborative efforts in dataset creation and model development hold promise for substantial advancements in the field.

\backmatter





\bmhead{Acknowledgements}

The authors would like to thank Van Lang University, Vietnam, for funding this work.

\section*{Statements and Declarations}

\begin{itemize}
\item \textbf{Competing Interests} The authors declare that there is no conflict of interest regarding the publication of this paper.
\item \textbf{Ethics approval} This study does not violate and does not involve moral and ethical statement.
\item \textbf{Consent for publication} The authors were aware of the publication of the paper and agreed to its publication.
\item \textbf{Data availability statement} The authors declare that data sharing not applicable to this article as no datasets were generated or analyzed during the current study.
\end{itemize}

\bibliography{sn-bibliography}

\end{document}